\documentclass{article}
\usepackage{microtype}
\usepackage{graphicx}
\usepackage{subcaption}
\usepackage{booktabs}
\usepackage{hyperref}
\usepackage{bm}
\usepackage{wrapfig}
\usepackage{xcolor}
\usepackage{amsmath}
\usepackage{enumitem}
\usepackage{comment}
\usepackage{url}

\newcommand{\e}{\mathrm{e}}

\DeclareMathOperator*{\softmax}{softmax}
\newcommand{\SF}[1]{\mathsf{#1}}
\newcommand{\DKL}{D_\mathrm{KL}}

\usepackage[preprint]{icml2026}

\usepackage{amsmath}
\usepackage{amssymb}
\usepackage{mathtools}
\usepackage{amsthm}

\usepackage[capitalize,noabbrev]{cleveref}

\theoremstyle{plain}

\theoremstyle{definition}

\theoremstyle{remark}

\usepackage[textsize=tiny]{todonotes}

\icmltitlerunning{Biased Generalization in Diffusion Models}

\begin{document}

\twocolumn[
  \icmltitle{Biased Generalization in Diffusion Models}
  \icmlsetsymbol{equal}{*}

  \begin{icmlauthorlist}
    \icmlauthor{Jérôme Garnier-Brun}{equal,B}
    \icmlauthor{Luca Biggio}{equal,B}
    \icmlauthor{Davide Beltrame}{B}
    \icmlauthor{Marc Mézard}{B}
    \icmlauthor{Luca Saglietti}{B}
  \end{icmlauthorlist}

  \icmlaffiliation{B}{Department of Computing Sciences, Bocconi University, Milan, Italy}

  \icmlcorrespondingauthor{Jérôme Garnier-Brun}{jerome.garnier@unibocconi.it}

  \icmlkeywords{Machine Learning, ICML}

  \vskip 0.3in
]

\printAffiliationsAndNotice{\icmlEqualContribution}

 \begin{abstract} 
    Generalization in generative modeling is defined as the ability to learn an underlying distribution from a finite dataset and produce novel samples, with evaluation largely driven by held-out performance and perceived sample quality.
    In practice, training is often stopped at the minimum of the test loss, taken as an operational indicator of generalization.    
    We challenge this viewpoint by identifying a phase of \emph{biased generalization} during training, in which the model continues to decrease the test loss while favoring samples with anomalously high proximity to training data. 
    By training the same network on two disjoint datasets and comparing the mutual distances of generated samples and their similarity to training data, we introduce a quantitative measure of bias and demonstrate its presence on real images. 
    We then study the mechanism of bias, using a controlled hierarchical data model where access to exact scores and ground-truth statistics allows us to precisely characterize its onset. 
    We attribute this phenomenon to the sequential nature of feature learning in deep networks, where coarse structure is learned early in a data-independent manner, while finer features are resolved later in a way that increasingly depends on individual training samples. Our results show that early stopping at the test loss minimum, while optimal under standard generalization criteria, may be insufficient for privacy-critical applications.
  \end{abstract}
\section{Introduction}

Generative AI can now produce text, images, and videos with a level of realism that was hardly imaginable only a few years ago, an achievement that also introduces profound societal challenges \cite{weidinger2023sociotechnical}. Whatever the medium, two questions stand at the center of current research: (i) does the generated content possess sufficient \emph{quality} to appear authentic, and (ii) is it truly \emph{novel} rather than a near-duplicate or patchwork of examples from the training set? More precisely, considering the task of learning to generate from a target distribution $P_0: \mathbb{R}^d \to \mathbb{R}$ given $n$ fair samples $\{\bm{x}^\mu\}_{\mu = 1,\dots,n}$, one should ensure that the generative process (i) samples according to a distribution $\tilde{P}^\theta_0$ that has a small distance to $P_0$, i.e., appears to achieve genuine \textit{generalization}; and (ii) does not generate individual samples $\bm{x}$ that are anomalously close to one of the training points $\{\bm{x}^\mu\}$, displaying some form of \textit{memorization}.

\begin{figure*}
    \centering
    \includegraphics[width=\linewidth]{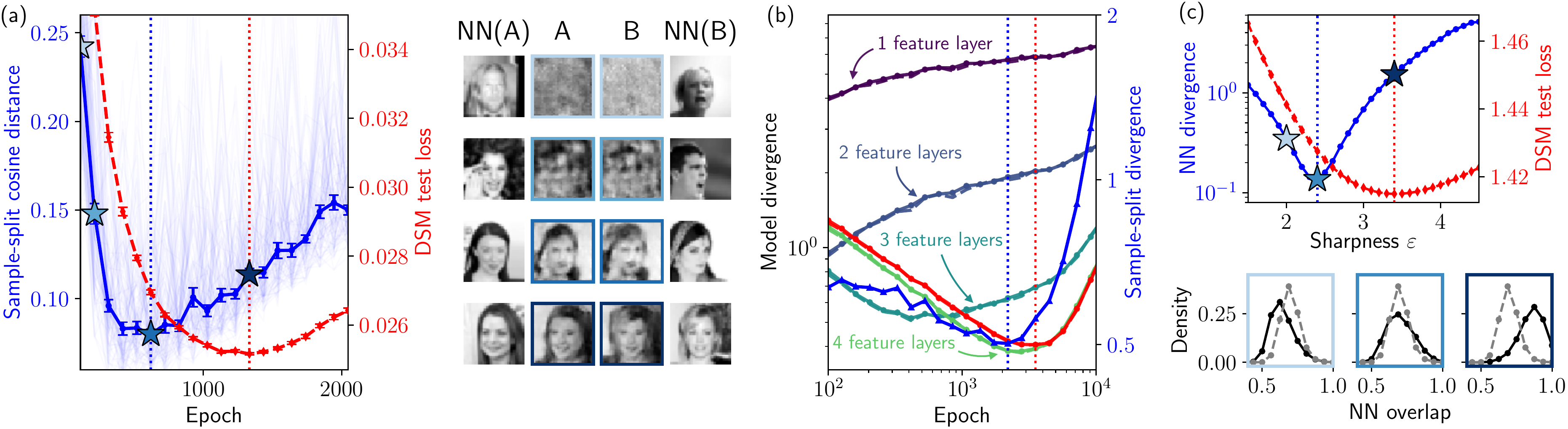}
    \vspace{-1.4em}
    \caption{
    Biased generalization emerges before overfitting across models and settings. (a) Sample-split analysis on CelebA: we compare two denoising diffusion models trained on non-overlapping data slices. Left: cosine distance between generated samples (left axis) and denoising score matching (DSM) test loss (right axis) during training, means over 15 models with standard errors. Generated samples become maximally similar before the test loss is minimal, indicating the onset of \emph{biased generalization} while the test loss is still decreasing. Colored stars mark epochs selected for visualization. Right: samples generated at the starred epochs by a model trained on each of the database split A/B (central columns), with nearest neighbors (NN) from each training split (side columns). Early in training, both models evolve similarly and sample quality improves; near the test-loss minimum, generated samples can differ substantially across splits and may get close to training examples, showing a bias without exact memorization. (b) Neural network trained on a controlled hierarchical dataset. Model-oracle divergence (left axis), representing the distance of the learned score to the exact score (red) and four coarser versions of the latter that account for lower-level features (as indicated by arrows). Sample-split analysis (right axis, blue) measuring the distance between denoising scores of models trained on disjoint datasets.  All scores are computed on test samples (w.r.t the model's training data), noised up to a critical diffusion time $t/T = 0.15$. The biased generalization phase is seen between the minimum of the sample-split curve (blue) and the minimum of the model-exact oracle divergence (red). It takes place when the models start resolving finer-scale features (light green). 
    (c) Training-free score model on the same hierarchical data, parametrized by a sharpness parameter $\varepsilon$ controlling the concentration of probability mass around the training data. Top: divergence between generated data and ground truth distributions of distances to the nearest neighbor (NN) in the training set (left axis) and DSM test loss (right axis) as a function of the sharpness parameter $\varepsilon$, showing sizeable bias at the test-loss minimum. Stars indicate selected values of $\varepsilon$. Bottom: in-training NN overlap (1 - normalized distance) distributions at selected sharpness levels, comparing model samples (solid black) to the ground truth (dashed gray). 
    }
    \label{fig:summary}
    \vspace{-1em}
\end{figure*}

The interplay between generalization and memorization is particularly relevant in the context of generative diffusion \citep{sohl2015deep}. There, as neural networks are typically trained to denoise a finite number of training samples, the minimum training loss is necessarily achieved by memorizing training examples \citep{gu2023memorization}. The prevailing view is that diffusion models generalize when they fail to memorize \citep{yoon2023diffusion}, whether due to limited architectural capacity \citep{george2025denoising} or favorable inductive biases \citep{kadkhodaie2023generalization,kamb2024analytic}. Recent work extends this picture to the training dynamics: diffusion models generalize on the way to memorization, with a plateau of good generative performance preceding the onset of overfitting \citep{bonnaire2025diffusion,favero2025bigger}; see also \citet{montanari2025dynamical} in a different context. This motivates early stopping at the minimum of the test loss, mirroring standard practice in supervised learning.

Nonetheless, as demonstrated by \citet{carlini2023extracting}, even models that appear to generalize well and are reported not to overfit in training, such as Imagen \citep{saharia2022photorealistic}, sometimes reproduce training samples almost exactly. This apparent paradox suggests that the dichotomy between generalization and memorization may be too coarse. 

The \textit{denoising score-matching} (DSM)  loss used to train diffusion models can be shown to be an upper bound of the Kullback-Leibler divergence between the true data distribution $P_0$ and the trained model distribution $\tilde{P}_0^\theta$ \citep{song2021maximum}. However, minimizing this loss does not preclude biased generation towards training examples. This disconnect, which was already identified by \citet{van2021memorization} in the context of variational autoencoders, highlights the need to move beyond aggregate generalization metrics and examine more localized signs of memorization or bias towards training-data in generative diffusion. Understanding the crossover between training-data-independent generalization and inevitable overfitting is essential to ensure that trained models are sufficiently accurate while not violating privacy or copyright-related constraints in relation to the training data.

In this paper, we tackle this issue in denoising diffusion probabilistic models (DDPM) \citep{ho2020denoising}. Prior efforts to understand the transition from generalization to memorization have focused on analytically tractable but simplified models \citep{li2023generalization,george2025denoising}, or on empirical studies using real images \citep{gu2023memorization,ross2024geometric}, the more detailed understanding being obtained by a mixture of these two strategies
\citep{bonnaire2025diffusion}, see also \citep{favero2025bigger}. Here, we study biased generalization using a similar two-pronged approach. We first demonstrate the emergence of bias before overfitting in models trained on real images. We then study the phenomenon in a controlled hierarchical data model with tunable long-range correlations, allowing us to precisely characterize the onset of bias thanks to the availability of ground-truth quantities.

\vspace{-0.3cm}
\paragraph{Contributions.} The key takeaway of our paper is:
\textit{In diffusion models, generalization and memorization can coexist, behaving as orthogonal---rather than opposite---axes of generative behavior.} More precisely:
\begin{itemize}[itemsep=-0.5pt,topsep=-2pt]
    \item We show that diffusion models trained on real images (CelebA) exhibit a phase of \emph{biased generalization}: two models trained on disjoint datasets start producing increasingly different outputs before either shows signs of overfitting, as measured by an increasing test loss, reflecting a growing bias toward their respective training samples (Fig.~\ref{fig:summary}(a)).
    \item We refine these findings in a controlled hierarchical data model, where trained denoisers can be analyzed with access to exact scores and ground-truth statistics, allowing us to confirm the emergence of bias through multiple complementary diagnostics (Fig.~\ref{fig:summary}(b)). 
    \item We provide a mechanistic account of why bias can emerge before overfitting, connecting the phenomenon to the sequential nature of feature learning in deep networks: while coarse structure is learned in a data-independent manner, the extraction of finer features becomes increasingly reliant on the available samples, yet generalization can continue to improve (Fig.~\ref{fig:summary}(b)).
    \item We illustrate the early onset of bias in a training-free simple score model on our hierarchical data, showing that \emph{biased generalization} is not an artifact of the inductive bias of neural networks or of SGD-based optimization dynamics (Fig.~\ref{fig:summary}(c)).
\end{itemize}

Section~\ref{sec:setup_sec} introduces the diffusion framework and the metrics used to detect bias, while the results are presented starting from Sec.~\ref{sec:biased_CelebA}.

\section{Diffusion models and bias}
\label{sec:setup_sec}

\subsection{Generative diffusion framework}
\label{sec:setup_generative_model}

Although the phenomenon of biased generalization is more general, we focus on denoising diffusion probabilistic models (DDPM) \citep{ho2020denoising}. In this framework, the forward process gradually corrupts data by adding Gaussian noise: at timestep $0 \leq t \leq T$, a clean sample $\bm{x}_0$ becomes
\begin{equation} \label{eq:forward}
\bm{x}_t = \sqrt{\overline{\alpha}_t} \bm{x}_0 + \sqrt{1-\overline{\alpha}_t}\bm{\xi}_t, \qquad \bm{\xi}_t \sim \mathcal{N}(0,\mathbf{I}_d)
\end{equation}
where $\overline{\alpha}_t$ controls the noise level, monotonically decreasing from $\overline{\alpha}_0 = 1$ to $\overline{\alpha}_T \approx 0$. Generation proceeds by reversing this process, which requires estimating the \textit{posterior mean}
\begin{equation}
\hat{\bm{x}}_0(\bm{x}_t) = \mathbb{E}[\bm{x}_0 \mid \bm{x}_t]
\end{equation}
at each noise level, which we will losely refer to as the \textit{score}. A neural network is trained to minimize the DSM loss on corrupted training samples, leading to an approximate posterior mean $\hat{\bm{x}}^\theta_0$. 

\vspace{-0.3cm}
\paragraph{Generalization and memorization.}
During training, supervision is local (restricted to noisy neighborhoods of the training samples in input space) yet most generative trajectories spend the majority of their evolution outside these regions \citep{song2025selectiveunderfittingdiffusionmodels}. The model’s ability to denoise unseen inputs, measured by the test loss, thus reflects underlying distributional structure and implies generalization. This notion, however, is in tension with the training objective. When training on a finite dataset $\{\bm{x}_0^\mu\}_{\mu=1,\ldots,n}$, the denoising score-matching (DSM) training loss is minimized by the \emph{empirical} denoiser, corresponding to the posterior mean computed with respect to the empirical data distribution concentrated on the training samples. As noted by \citet{gu2023memorization}, a sufficiently expressive network perfectly minimizing the training loss will therefore eventually memorize the training set. Early stopping, i.e. halting training at the minimum of the test loss, is commonly invoked to mitigate this tension, see e.g. \citep{li2023generalization}, as it can formally be shown that the test DSM loss is an upper bound to the $\DKL(P_0 \mid\mid \tilde{P}_0^\theta)$ \cite{song2021maximum}.

\vspace{-0.3cm}
\paragraph{Detecting memorization.}

A common approach for detecting memorization is to compare generated samples to their nearest neighbors in the training set. For instance, on real datasets, \citet{yoon2023diffusion} define a generated sample $\bm{x}$ as \emph{replicating} a training point if
\begin{equation} \label{eq:replicating}
    \frac{\| \bm{x} - \bm{x}^a_0\|_2}{\| \bm{x} - \bm{x}^b_0\|_2} < \frac{1}{3},
\end{equation}
where $\bm{x}_0^a$ and $\bm{x}_0^b$ denote the closest and second-closest samples in the training dataset under the $\ell_2$ norm. 
Empirically, such proximity-based indicators are found to become non-negligible only after the test loss starts increasing, i.e., in the overfitting regime. This observation supports the widespread interpretation of memorization as a failure mode that emerges in opposition to generalization. In high-dimensional settings, these two generative regimes have been found to occur on different time scales 
\citep{bonnaire2025diffusion,favero2025bigger} separated by a long plateau of good generalizing behavior. By contrast, we find that on structured finite-dimensional data, the generalization phase can itself be subdivided into a universal---sample independent---and a biased phase.

\subsection{Measuring bias in practice}
\label{sec:bias_metrics}
We define \emph{bias} in a generative model as the emergence of anomalous similarity between generated samples and the training data, relative to what would be expected under i.i.d. sampling from the underlying data distribution. Unlike overt memorization, such bias can be subtle and may arise well before individual training samples are nearly reproduced. Nevertheless, this notion is relevant in settings where data privacy is critical, and the systematic reuse of training-specific features may be undesirable even in the absence of exact sample replication. Detecting bias therefore requires a reference notion of unbiased behavior, against which model outputs can be compared. Bias can be probed through complementary observables, either at the level of generated samples or at the level of the learned denoising function.

\vspace{-0.3cm}
\paragraph{Sample-level.}
Rather than relying on binary memorization criteria, we consider the full distribution of distances between generated samples and their nearest neighbors in the training set. Along training, the generated data should shift from being anomalously far (random initialization) to anomalously close (eventual memorization) to training samples, see Fig.~\ref{fig:summary}(c). By analyzing the Kullback-Leibler (KL) divergence from the unbiased distribution---which we term the \emph{nearest-neighbor divergence}---one can detect systematic deviations toward smaller distances, indicating biased generation toward the training data. Alternatively, one can also compare the samples generated by the model and the reference in identically seeded reverse diffusion trajectories \citep{kadkhodaie2023generalization}, thereby isolating the impact of the denoising functions.

\vspace{-0.3cm}
\paragraph{Score-level.}
We propose to measure how much the trained and reference denoising scores differ, when evaluated on the same noisy inputs at fixed diffusion times. This time-resolved analysis allows one to localize the onset of bias along the diffusion trajectory and to focus on regimes where denoising is strongly input-dependent. In addition, it enables conditioning on specific starting points of the backward process, thereby concentrating the comparison on regions where biasing effects are most pronounced.
 
In practice, the reference against which these observables are evaluated depends on the available information. In controlled settings, such as synthetic data models, one may have direct access to the ground-truth distribution or to exact scores, enabling direct comparison between model outputs and oracle quantities. In realistic settings with finite data, we use a surrogate reference obtained through a \emph{sample-split analysis}: two models are trained separately on independent subsets of data \citep{kadkhodaie2023generalization}. If both models remain in an unbiased regime, their predictions are expected to converge toward the same population-level behavior \citep{favero2025bigger}. Growing discrepancies between the models signal data-dependent bias, even if the test losses continue to improve.

\section{Experiments on real data}
\label{sec:biased_CelebA}

\paragraph{Experimental setup.} 

We follow the numerical setup of \citet{bonnaire2025diffusion} on the CelebA face dataset \citep{liu2015celeba}, which we convert to grey-scale downsampled images of size $32\times 32$.
The generative process is the DDPM described above, with a U-Net architecture \citep{ronneberger2015unet} trained to predict additive Gaussian noise $\hat{\bm{\xi}}_t$ from the noised input $\bm{x}_t$. We provide implementation details in Appendix~\ref{appendix:celeba_exps}.

\begin{figure}
    \centering
    \includegraphics[width=\linewidth]{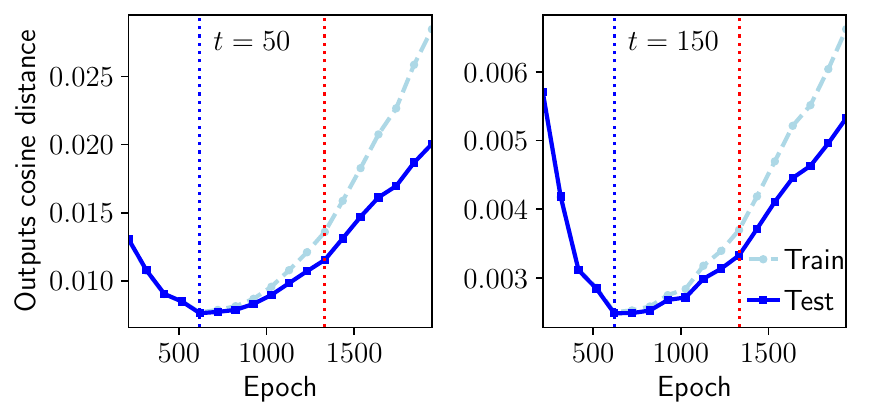}
    \vspace{-1.8em}
    \caption{Cosine distance between the predictions of two networks trained on disjoint subsets of CelebA of size $n = 1024$, evaluated on inputs noised until time $t$ out of $T = 1000$. The distance is evaluated for original images that are either outside of both training sets (``Test'') or in \emph{one of them} (``Train''). The vertical dashed lines correspond to the minima of the diffusion time-averaged metrics shown in Fig.~\ref{fig:summary}(a).}
    \label{fig:CelebA_scores}
    \vspace{-1em}
\end{figure}

\vspace{-0.3cm}
\paragraph{Results.}
We first measure bias at the sample level. Following identical noise trajectories, we generate samples from models trained on distinct datasets of size $n=1024$ and quantify their divergence using the cosine distance $D_C = 1 - S_C$, where $S_C(\bm{x},\bm{y}) = \frac{\bm{x}\cdot\bm{y}}{\|\bm{x}\|\,\|\bm{y}\|}$. This \emph{sample-split analysis}, averaged over all pairings between 15 trained models, is shown in Fig.~\ref{fig:summary}(a). Despite substantial variability across runs (light shaded curves), the mean exhibits a clear U-shape, with a minimum reached early in training. The DSM test loss, also shown in Fig.~\ref{fig:summary}(a), follows a similar trend but reaches its minimum significantly later. This separation precisely identifies the \emph{biased generalization} phase. 
Note that the standard indicator of memorization defined in Eq.\,\eqref{eq:replicating} records a maximum average value across checkpoints below $10^{-3}$.

We illustrate the bias effect qualitatively using representative generated samples in Fig.~\ref{fig:summary}(a) (right; see Appendix~\ref{appendix:celeba_exps}. for details). Up to the maximum similarity point (third row of samples), the two models generate nearly identical images (central columns). Near the minimum of the test loss, the samples develop marked differences and can exhibit features closely resembling their respective nearest training examples (side columns), without exact replication.  

We further validate this behavior at the score level. Fig.~\ref{fig:CelebA_scores} shows the cosine distance between denoiser outputs of two models evaluated at fixed diffusion times, with the minima of the sample-level bias metric and of the test loss indicated by vertical dashed lines. 
Importantly, this normalized metric is able to quantify a misalignment between the two models already during the early transient in training, when both produce outputs of small magnitude and the generative process is almost entirely driven by noise.
Focusing on short ($t=50$) and intermediate ($t=150$) diffusion times, where denoising is most input-dependent, we observe that score divergence starts increasing close to the onset of sample-level bias, again well before the test loss minimum. Additional evidence is obtained by conditioning on the origin of the noisy input: while the score distance is independent of whether inputs originate from the training or test sets in the unbiased phase, a clear separation emerges after the bias minimum. At the test-loss minimum, the gap between training- and test-conditioned curves confirms the presence of training-data bias in the learned denoisers.

\section{Biased generalization in a controlled setting}
\label{sec:biased_controlled}

\subsection{Data model}
\label{sec:data_model}

To verify and better understand the biased generalization that we have observed on real images, we now consider a controlled hierarchical data setting with explicit compositional structure, inspired by recent work on context-free grammars \citep{zhao2023transformers,allen2023physics, garnier2024transformers,cagnetta2024deep}. In particular, we generate discrete sequences, $\bm{s}^\mu \in \{1,\dots,q\}^N$ via a tree-based graphical model, characterized by unambiguous production rules $a\to bc$ with sparse, log-normally distributed weights (more details in Appendix~\ref{appendix:rhm_exps}).
The generation process is repeated over $\ell$ layers, leading to sequences of size $N = 2^\ell$. Unless indicated otherwise, we use $\ell = 4$ layers and $q = 6$, see Appendix~\ref{appendix:rhm_exps} for further details.

To apply the formalism of continuous diffusion to discrete data, we follow \citet{li2022diffusion} and one-hot encode the sequences $\bm{x}_0^\mu=\mathrm{onehot}_q(\bm{s}^\mu)$, where $\bm{x}_0^\mu\in\mathbb{R}^d$ and $d = N q$.
Note that this implies that the output of a trained denoiser approximating the true posterior mean $\mathbb{E}(\bm{x}_0 \mid \bm{x}_t)$ can be interpreted as a normalized \textit{marginal probability distribution} over the associated discrete symbol. As a result, the metric of choice to compare posterior means in our data model is the Kullback-Leibler divergence $D_{\mathrm{KL}}$, rather than the cosine distance between noise vectors used in Sec.~\ref{sec:biased_CelebA}.

\vspace{-0.3cm}
\paragraph{Exact inference and hierarchical filtering.}
The tree structure of the data generative model, sketched in Fig.~\ref{fig:reverse_BP}(a), has two major advantages compared to real data. Firstly, it enables straightforward i.i.d.\ sampling through a Markov chain initialized at the root of the tree. Secondly, it allows exact inference of the posterior mean given a noisy observation, $\hat{\bm{x}}_0(\bm{x}_t)$, via a dynamic programming algorithm known as Belief Propagation (BP) \cite{mezard2009information}. Further details on the associated message-passing equations and conditioning on the observation are provided in Appendix~\ref{appendix:rhm_exps}. 
In addition, the hierarchical structure of the model makes it possible to perform controlled filtering of correlations, following the procedure introduced by \citet{garnier2024transformers}. By treating the nodes at an intermediate layer of the tree, say at level $k<\ell$, as conditionally independent given the root (Fig.~\ref{fig:reverse_BP}), one can selectively suppress long-range correlations in the data. In particular, high-level features involving contiguous blocks of symbols of size $2^{\ell-k}$ or greater are filtered out. The exact BP algorithms associated with these filtered topologies, denoted BP$_k$, can be interpreted as performing increasingly coarse inference over the original hierarchical data. These tools will be used to characterize the emergence of structural scales along the reverse diffusion process and to identify which features are learned by a trained model. Additional details are provided in Appendix~\ref{appendix:rhm_exps}.

\vspace{-0.3cm}
\paragraph{Experimental setup.} Building on the findings of \citet{garnier2024transformers}, we adopt a vanilla transformer encoder architecture  \citep{vaswani2017attention} for the denoiser, which is able to implement a close approximation of BP in this data model. 
In our context, the model output is transformed via a $q$-wise softmax, allowing $\hat{\bm{x}}_0^\theta(\bm{x}_t)$ to be interpreted as a marginal probability distribution over the discrete symbols. We therefore train the model by minimizing the cross-entropy loss between these normalized outputs and the one-hot encoded noise-free original training samples. We use $T=500$ diffusion steps and a standard linear noise schedule. More details of our implementation and training algorithm can be found in Appendix~\ref{appendix:rhm_exps}.

\subsection{Dynamical emergence of bias}
\label{sec:exact_phenomenology}

\begin{figure}  \centering
  \includegraphics[width=\linewidth]{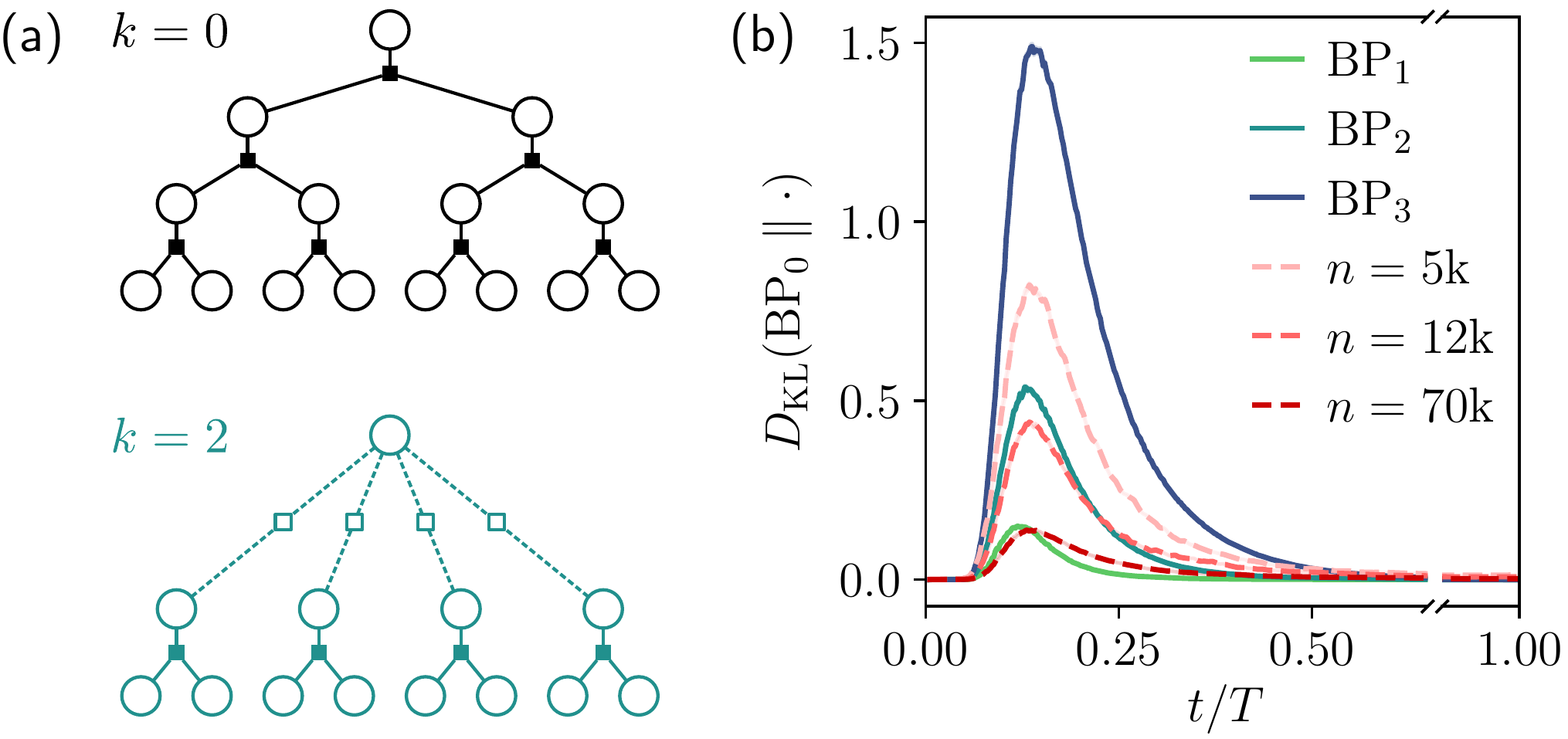}
  \caption{(a) Illustration of an $\ell = 3$ tree-based data generation for the full model (top) and a hierarchically filtered version with $k=2$ (bottom). (b) Kullback-Leibler divergence between the exact posterior mean of BP$_0$ and those from filtered BP$_k$ denoisers along a reverse trajectory following the exact BP, averaged over $2\mathrm{k}$ realizations. Dashed lines show trained models at the minimum of their test losses for different training set size $n$.}
    \label{fig:reverse_BP}
    \vspace{-1em}
\end{figure}

We focus our score-level comparisons on regions of diffusion time where reverse trajectories produced by different denoisers are most sensitive to the structural properties of the data, and where bias can therefore emerge most clearly. Following \citet{biroli2024dynamical}, we can delineate distinct dynamical regimes along the reverse diffusion process. In Fig.~\ref{fig:reverse_BP}(b), we report the Kullback--Leibler divergence between the outputs of filtered (BP$_k$) and trained denoisers with the full-tree oracle BP$_0$, evaluated on noisy observations of i.i.d. data. Three regimes can be identified:

\emph{(i) Long-time unstructured regime.}
For large diffusion times, $t/T \gtrsim 0.5$, the signal contained in $\bm{x}_t$ is negligible and all denoisers output the unconditional marginal probabilities $\mathbb{P}(\bm{x}_0)$. As a result, the filtered and trained denoisers are nearly indistinguishable from the exact oracle, and no sign of bias can be detected. This long-time \emph{unstructured regime} is analogous to phase~I identified in \citet{biroli2024dynamical}.

\emph{(ii) Intermediate structured regime.}
At intermediate times, $0.08 \lesssim t/T \lesssim 0.5$, the outputs of BP$_k$ and of trained denoisers progressively depart from the exact BP$_0$ posterior, reflecting the increasing relevance of longer-range correlations in $\bm{x}_t$ for the oracle prediction. In line with \citet{garnier2024transformers}, increasing the training set size allows trained denoisers to resolve progressively finer structure (pink--red dashed lines) and decrease the distance to the oracle. 
For all imperfect denoisers, the KL distance to the exact posterior peaks around a critical time $t/T \approx 0.15$, consistent with the observations of \citet{sclocchiprobing} in discrete diffusion models. We therefore select this critical point for fixed-time score comparisons (Fig.~\ref{fig:summary}(b), Fig.~\ref{fig:sequential}).

\emph{(iii) Short-time trivial regime.}
At short diffusion times, $t/T \lesssim 0.08$, all denoisers again appear effectively equivalent. In this regime, the noise level is sufficiently low that denoising i.i.d. sequences can be achieved by simple symbol-wise rounding, without exploiting any correlation structure, a behavior specific to the discrete data setting. Note that differences between denoisers would instead emerge when considering neighborhoods of sequences that are out-of-distribution for the oracle score but remain admissible for models that ignore long-range correlations. However, because the supervision signal during training is restricted to weakly noised versions of valid training samples, it provides no information that would distinguish the exact (or empirical) score from such a naive symbol-wise prior in this regime. As a result, even imperfectly trained denoisers converge to the simplest explanation compatible with the loss and exhibit identical behavior at short times (see Fig. \ref{fig:Uturn_randomstart} in Appendix \ref{appendix:extra_figs}), rendering this regime uninformative for probing bias. 

\subsection{Early-stopped models are biased} \label{sec:biased_models}

\begin{figure}
    \centering
    \includegraphics[width=\linewidth]{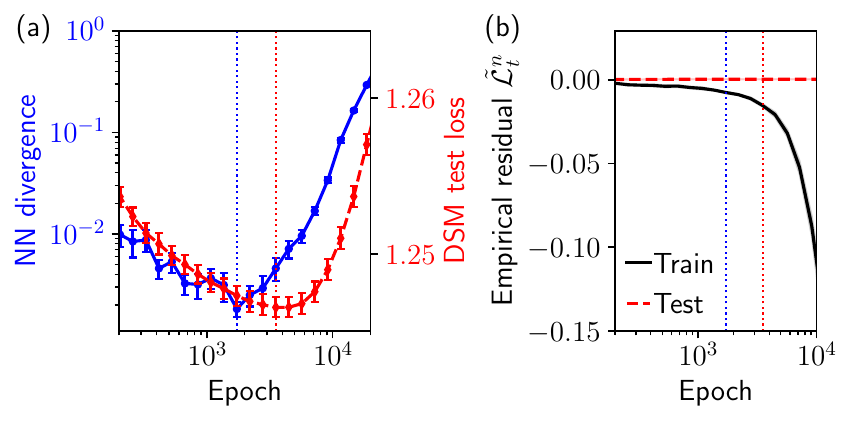}
    \vspace{-1.8em}
    \caption{Bias metrics for a diffusion model trained on $n= 12\mathrm{k}$ tree-based sequences computed with 50$\mathrm{k}$ generated samples and evaluation points, and averaged over 15 training runs and showing the standard error. (a) Nearest-neighbor divergence (Sec.~\ref{sec:bias_metrics}) of generated sequences and denoising test loss as a function of training.  (b) 
    Expectation value of the excess data-dependent loss of (4) 
    evaluated on test and train sequences noised up to time $t = 150$. Vertical dashed lines show the minimum of the bias metric (blue) and test loss (red).}
    \label{fig:Hierarchical_bias_evidence}
    \vspace{-1em}
\end{figure}

\paragraph{Sample-split analysis.}
As in Sec.~\ref{sec:biased_CelebA}, we independently train two denoisers on disjoint subsets of the data and monitor the model divergence, quantified by the $D_{\mathrm{KL}}$ between their predicted posterior means on identical noised inputs. The key question is when this divergence emerges relative to standard generalization metrics. 
In the present controlled setting, access to the exact score via BP allows us to additionally track the $D_{\mathrm{KL}}$ between each trained model and the oracle posterior mean. The minimum of this quantity marks the point of closest approximation to the true posterior and provides a sample-efficient estimate of the test-loss minimum (see loss decomposition below). As shown in Fig.~\ref{fig:summary}(c), the onset of model divergence occurs \emph{before} this minimum: the models enter a data-dependent regime while still improving toward the exact score, thereby confirming the existence of a biased generalization phase.

\vspace{-0.3cm}
\paragraph{Nearest-neighbor divergence.}
We next examine the distribution of nearest-neighbor overlaps---here the fraction of symbols in agreement between two sequences--between generated samples and the training data, taking advantage of our ability to sample i.i.d. from our data model. Fig.~\ref{fig:Hierarchical_bias_evidence}(a) shows the nearest-neighbor divergence as a function of training, displaying a minimum reached significantly before the minimum of the DSM test loss. 
The training epoch at which this metric begins to increase closely matches the onset of model divergence identified by the sample-split analysis in Fig.~\ref{fig:summary}(c), confirming the consistency of the two diagnostics of generative bias. The reported curves are averaged over 15 independently trained models, as the adaptive optimizer can induce abrupt fluctuations in this extreme-value statistic along training; see Appendix~\ref{appendix:extra_figs_hierarchical}.

\vspace{-0.3cm}
\paragraph{Identifying bias with a loss decomposition.}
With access to the exact posterior mean, we derive and evaluate an explicit decomposition of the denoising loss that makes the buildup of data bias transparent. Each cross-entropy loss term can be rewritten as
\begin{equation}
\begin{aligned}
    \ell_t(\theta, \bm{x}_0,\bm{x}_t)
    = &\;\overbrace{- \hat{\bm{x}}_0(\bm{x}_t)^\top \log \hat{\bm{x}}^\theta_0(\bm{x}_t)}^{\text{Exact distillation } \ell^\star_t} \\
    &\;\;-\underbrace{(\bm{x}_0 - \hat{\bm{x}}_0(\bm{x}_t))^\top \log \hat{\bm{x}}^\theta_0(\bm{x}_t)}_{\text{Excess data-dependent } \tilde{\ell}_t},
\end{aligned}
\end{equation}
where the logarithm is taken element-wise. The test denoising loss is obtained by taking the expectation $\mathbb{E}_{\bm{x}_0, \bm{x}_t}[\ell_t(\theta, \bm{x}_0,\bm{x}_t)]$ over the joint distribution of $(\bm{x}_0,\bm{x}_t)$ induced by the draw of data and diffusion noise. For fixed model parameters $\theta$, we have $\mathbb{E}_{\bm{x}_0}\!\left[ \tilde{\ell}_t(\theta, \bm{x}_0,\bm{x}_t) \mid \bm{x}_t \right] = 0$, and therefore, by the law of total expectation, $\mathbb{E}_{\bm{x}_0, \bm{x}_t}[\tilde{\ell}_t] = 0$. As a consequence, the excess term, measuring data-dependent overconfidence, does not contribute on average to the test loss, as verified in  Fig.~\ref{fig:Hierarchical_bias_evidence}.
In contrast, for the training loss the expectation over $\bm{x}_0$ is replaced by the empirical average over the training set. In this case, the excess loss, quantifying the mis-calibration of the model relative to the exact score induced by the pull of the empirical distribution, is finite and can be optimized at training. As shown in Fig.~\ref{fig:Hierarchical_bias_evidence}(b), this term starts decreasing significantly in the biased-generalization phase. Crucially, before the minimum of the test loss, the model is able to simultaneously optimize both the distillation term and the excess data-dependent term, providing a concrete illustration of orthogonality and coexistence of generalization and memorization.

\begin{figure}
  \centering
  \includegraphics[width=\linewidth]{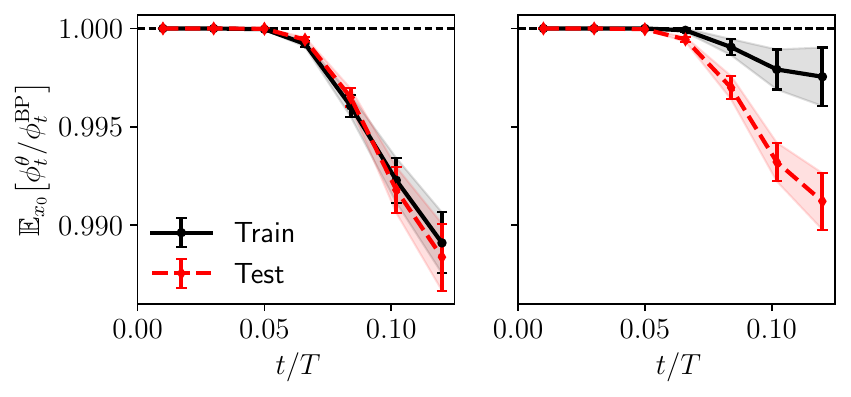}
\vspace{-1.8em}
  
  \caption{Average normalized overlap to the starting sequence for a ``U-turn'' experiment after noising to time $t$ using $100$ trajectories for each $1000$ starting points, model trained on $n = 12\mathrm{k}$ data. Left: checkpoint minimizing the Kullback-Leibler divergence of nearest neighbor overlaps with the result expected from fair sampling. Right: checkpoint minimizing the denoising test loss. Shaded areas show the standard error over starting sequences.}
    \label{fig:model_uturn}
    \vspace{-1em}
\end{figure}

\vspace{-0.3cm}
\paragraph{Bias in the reverse process.} We further probe the \emph{dynamical} emergence of bias along diffusion trajectories by means of a ``U-turn'' experiment. Starting from a clean sample $\bm{x}_0 \sim P_0$, we noise it up to diffusion time $t$. We then follow the reverse \textit{process} using a given denoising score, yielding a reconstructed sample $\tilde{\bm{x}}_0$ at $t=0$. Bias can be detected by comparing the outcomes of such U-turns when conditioned on starting points drawn from the training set versus previously unseen data, using BP as a reference for unbiased behavior.

For each fixed starting point $\bm{x}_0$, we compute the overlap between $\tilde{\bm{x}}_0$ and $\bm{x}_0$ over multiple U-turn realizations, using either the trained model or BP as the denoising score. This defines the average overlaps $\phi_t^\theta(\bm{x}_0)$ and $\phi_t^{\mathrm{BP}}(\bm{x}_0)$. We then consider the ratio $\phi_t^\theta(\bm{x}_0)/\phi_t^{\mathrm{BP}}(\bm{x}_0)$, which isolates deviations of the trained model from optimal (oracle) behavior, and finally average this quantity over starting sequences. 

Figure~\ref{fig:model_uturn} shows the evolution of $\mathbb{E}_{\bm{x}_0}[\phi_t^\theta/\phi_t^{\mathrm{BP}}]$ for a trained model ($n=12\mathrm{k}$) at the checkpoint minimizing the nearest-neighbor divergence and at a later checkpoint minimizing the test loss. At the former, the curves corresponding to training and test starting points are indistinguishable within statistical fluctuations. At the test-loss minimum, by contrast, a clear separation emerges: corrupted training samples are significantly more likely to be recovered by the reverse process than previously unseen data, providing direct dynamical evidence of training-data bias. At the same time, both curves move closer to unity, corresponding to oracle performance, even on the test set. This illustrates once again how the test loss can continue to decrease despite the emergence of bias.\footnote{The ratio remains below one because the trained model only partially approximates the BP denoiser; see Sec.~\ref{sec:sequential_bias}.} We show a more complete evolution of such U-turn experiments along training and for another trained model in Appendix~\ref{appendix:extra_figs_hierarchical}.

\section{Understanding the emergence of bias}

\subsection{Sequential learning and bias}
\label{sec:sequential_bias}

Both in real data and in the hierarchical data setting, we have shown that \emph{biased generalization} does not emerge immediately during training, nor exactly at the test loss minimum. We now discuss the underlying mechanism, which can be understood in detail in our controlled setting, refining the description provided in \citep{favero2025bigger}.

\begin{figure}
    \centering
    \includegraphics[width=\linewidth]{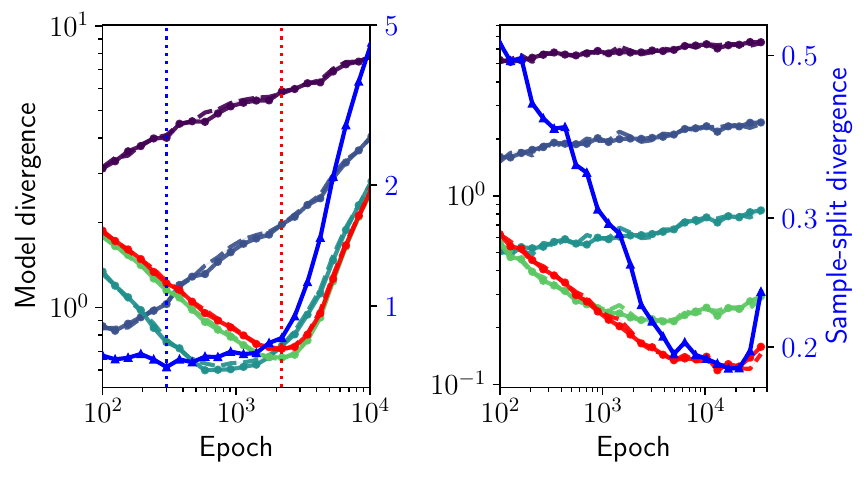}
    \vspace{-1.8em}
    \caption{Evolution of model divergences $\DKL(\cdot \mid\mid \hat{\bm{x}}_0^\theta)$ for different BP$_{k}$ (from purple to green, decreasing in $k$) and BP$_0$ (red), evaluated on models A ($-\bullet$) and B ($--$). Blue curves ($-\blacktriangle$) show the sample-split denoising score divergence $\DKL(\hat{\bm{x}}_0^{\theta_A} \mid\mid \hat{\bm{x}}_0^{\theta_B})$. Left: $n = 5\mathrm{k}$, Right: $n=70\mathrm{k}$.}
    \label{fig:sequential}
    \vspace{-1em}
\end{figure}

\vspace{-0.3cm}
\paragraph{Bias mechanism.} In the context of tree-based data models, the $\ell$ hierarchical levels of features to be learned induce a sequential, staircase-like discovery process \citep{garnier2024transformers,favero2025compositional}. On the one hand, for a given number of training samples $n$, the network has sufficient statistics to reliably learn only the first $\ell - k^\star(n)$ levels of the hierarchy, with $0 \leq k^\star(n) \leq \ell$. On the other hand, along the training dynamics, the model learns features in order of increasing complexity. While the model is focusing on simpler, shorter-range features corresponding to layers $k > k^\star(n)$, it follows a gradient signal that is effectively indistinguishable from that of the population loss in the early dynamics, exhibiting a ``mean-field'' feature learning regime \citep{montanari2025dynamical}. As long as the network remains in this \emph{partial generalization} regime \citep{favero2025bigger}, its representation is approximately unbiased, and the outputs of models trained on different data splits remain aligned.
If one trains the model further, the training data are insufficient to fully resolve the next level, so the models adopt a data-biased approximation of it, resulting in a simultaneous decrease in the test loss and an increase in any of the considered bias measurements. We have shown that the onset of this biased phase occurs before the minimum of the test loss.

\vspace{-0.3cm}
\paragraph{Empirical validation.} We verify this picture by leveraging the BP$_k$ descriptions (Sec.~\ref{sec:data_model}), i.e., oracle denoisers that resolve exactly $\ell - k$ levels of the hierarchy. In Fig.~\ref{fig:summary}(b), we show the $D_{\mathrm{KL}}$ between each BP$_k$ and the transformer denoiser along training, for an intermediate training set size $n = 12\mathrm{k}$. We contrast these curves with the sample-split score divergence, detecting the emergence of bias. As training progresses, the model sequentially approaches BP$_k$ with decreasing $k$, reflecting resolution of longer-range correlations. For this value of $n$, the best match is achieved with BP$_1$, indicating that the model cannot fully resolve the highest level of the hierarchy and thus $k^\star = 1$. Beyond the maximal similarity to BP$_1$, the model continues to improve towards the ground truth while bias starts increasing. In Fig.~\ref{fig:sequential}, we show analogous plots for smaller ($n = 5\mathrm{k}$) and larger ($n = 70\mathrm{k}$) training sets. With fewer samples, the best match occurs at a higher $k^\star = 2$, and the biased region is wider; with more samples, the model approaches BP$_0$, and the bias becomes negligible. Importantly, in all cases the onset of bias follows the epoch at which the best-matching BP$_{k^\star}$ is reached.  
We note that the sequential learning dynamics we describe is not perfectly rigid: the transitions between levels are gradual rather than sharp, particularly when using adaptive optimizers such as Adam. Nonetheless, the overall picture of progressive feature learning followed by data-dependent approximation of finer structure provides a coherent explanation for the emergence of biased generalization. 

\subsection{A training-free example of biased generalization}
\label{sec:toy_model}
The sequential discovery of features plays a key role in locating the biased generalization phase in our trained model, yet this phenomenon is not unique to neural network denoisers and can arise in much simpler settings. 
We now consider a single-parameter probability distribution focused on the training data, still sampled i.i.d. from the tree-generative model. The parameter $\varepsilon$, which we refer to as the \emph{sharpness} of the distribution, controls the concentration of the probability mass on the data, by assigning probability weight $\propto1$ to the correct symbols, and a uniform weight $\propto \e^{-\varepsilon}$ to all the others.  
The associated denoising score can thus be interpreted as a smoothed version of the empirical score---exactly recovered in the limit $\varepsilon\to\infty$, while the $\varepsilon\to0$ limit yields a uniform distribution. Full details are reported in Appendix~\ref{appendix:toy_model}.

We can use this parametric estimator $\tilde{P}^\varepsilon_0$ in generative diffusion by exactly computing the associated posterior means, and employing them in the reverse process. Thus, we can monitor the DSM test loss as a function of $\varepsilon$, while efficiently generating samples to detect bias.
In Fig.~\ref{fig:summary}(a), we plot the nearest-neighbor divergence alongside the test loss, as a function of the sharpness $\varepsilon$. 
Consistent with our findings on trained models, we detect the existence of a wide phase of \emph{biased generalization}, suggesting that a misalignment between distribution matching and sample fairness objectives might emerge beyond the specific settings analyzed in this paper. Note that a high-dimensional counterpart to this training-free example was analyzed in \citep{biroli2024kernel}, where the optimal width (that minimizing  the KL divergence with $P_0$) in a simple Kernel Density Estimation problem was located beyond a phase-transition 
where the shape of the estimator becomes strongly data-dependent. This exactly solvable example of biased generalization thereby demonstrates that the phenomenon may arise even in asymptotic regimes. 

\section{Limitations}

Our study focuses on the standard DDPM setting, and does not address whether analogous biased regimes arise with other classes of modern generative models. In the context of the discrete hierarchical model that we use, it would be particularly relevant to assess whether more specialized generative processes, such as discrete diffusion or other autoregressive models, could perhaps mitigate the effect. Similarly, while we provide evidence on real images and in a controlled setting, our empirical evaluation is also necessarily limited to moderate-scale datasets and architectures, and we do not explore the behavior of large-scale diffusion models trained on higher dimensional data.

Moreover, the purpose of our analysis is primarily diagnostic: we do not propose mitigation strategies or alternative training objectives designed to prevent it. Finally, bias is quantified through proximity-based and score-based metrics, which cannot capture semantic similarity or downstream perceptual notions of memorization. 

\section{Conclusion}

In this work, we identify and characterize a regime of \emph{biased generalization} in generative diffusion models, in which the test loss continues to decrease while the generative behavior of the model becomes increasingly dependent on the specific training samples. At a conceptual level, this phenomenon reflects a simple but important fact: minimizing a distributional discrepancy between a model and the data-generating process does not, by itself, preclude biased sampling toward the finite dataset used for training. While related observations had previously been made in other generative settings \citep{van2021memorization, biroli2024kernel}, we show that this effect arises naturally and universally in diffusion models.
By combining empirical evidence on real images with a controlled hierarchical data model, we trace the early emergence of bias to the sequential nature of feature learning in deep networks \citep{refinetti2023neural, bardone2025theory}, whereby increasingly fine structure is approximated in a data-dependent manner once available statistics become insufficient. As a result, biased generalization can arise \emph{before} the test loss minimum, highlighting a limitation of early stopping as a safeguard against training-data bias.
These observations have practical implications for the evaluation and deployment of generative models. Implicitly conflating generalization with unbiased sampling could become particularly consequential as generative models are deployed in domains where privacy, memorization, and data provenance are critical. Beyond the issue of training-data bias, defining appropriate metrics to quantify (or even clearly define) generalization remains a challenge in self-supervised learning, see e.g. \citet{mendes2026solvable}.

An important open direction is to understand how common generation mechanisms may interact with the biased generalization regime. In particular, conditioning and guidance techniques such as classifier-free guidance \citep{ho2022classifier} may amplify subtle data-dependent biases by selectively steering generation toward features overfitted to specific training data \citep{carlini2023extracting, somepalli2023diffusion}. Clarifying the interplay between biased generalization, conditioning, and data extraction is a promising direction for future work.

\section*{Acknowledgements}
The authors are grateful to G. Biroli, S. Moran, P.-F. Urbani and R. Urfin for stimulating discussions and to E. Moscato who participated in the early stages of this project. The work of J.\,G.-B. was supported by the European Union’s Horizon Europe program under the Marie Skłodowska-Curie grant agreement No.~101210798.

\bibliography{refs}
\bibliographystyle{icml2026}

\newpage
\appendix
\onecolumn

\section{Additional details on CelebA experiments}\label{appendix:celeba_exps}

\subsection{Experimental setup} 

Our experiments leverage the codebase of \citet{bonnaire2025diffusion}, available at \url{https://github.com/tbonnair/Why-Diffusion-Models-Don-t-Memorize}, and follow closely their setup, see \url{https://github.com/davide-beltrame/biased-generalization} for our slightly modified version.
\vspace{-0.2cm}
\paragraph{Dataset.}
We use the CelebA dataset \citep{liu2015celeba}, center-cropped, gray-scaled and resized to $32 \times 32$ pixels. No data augmentation is used. All the considered models are trained of different subsets of size $n=1024$ of the original dataset.

\paragraph{Neural network architecture and diffusion.}
We use a standard U-Net architecture \citep{ronneberger2015unet} with base channel width 32 and channel multipliers $(1,2,4)$. Self-attention is applied at the $16\times16$ and $8\times8$ resolution levels (i.e. the second and third encoder/decoder blocks).
The network is used within the standard DDPM \citep{ho2020denoising} framework and is trained to predict additive Gaussian noise across $T=1000$ steps with a linear noise schedule for $\beta_t$, ranging from $10^{-4}$ to $0.02$.

\paragraph{Training and generation.}
All models are trained with the Adam \citep{kingma2017adam} optimizer, with a learning rate $\eta = 1 \times{10^{-4}}$, $(\beta_{1},\beta_{2}) = (0.9,0.999)$ and a batch size of 512. Training time is measured in epochs, where within each epoch each training point is corrupted according to Eq. \eqref{eq:forward}, with a timestep $t$ drawn uniformly at random in $[1,1000]$. All our models are trained for 5000 epochs.

To generate images from a trained model, we follow the standard DDPM procedure: starting from white noise, we repeatedly compute:
\begin{align}
\bm{x}_{t-1}
&=  \frac{1}{\sqrt{\alpha_t}} \left(\bm{x}_t - \frac{1-\alpha_t}{\sqrt{1-\overline{\alpha}_t}} \widehat{\epsilon}_{\boldsymbol{\theta}}(\bm{x}_t)\right) + \sqrt{\beta_{t}} \bm{z}, \qquad \bm{z} \sim \mathcal{N}(0,\mathbf{I}).
\label{eq: DDPM reverse noise predict}
\end{align}
for $t = T, ..., 1$, where $\widehat{\boldsymbol{\epsilon}}_{\theta}(\bm{x}_t)$ is the output of the model given $\bm{x}_t$. 

\subsection{Score-level experiments}
As mentioned in the main, for CelebA we do not have access to the exact score. For this reason, we perform a sample-split analysis by comparing the scores of two models (A and B) trained on different subsets of the original dataset. We train both models for the same number of epochs and compare them at corresponding training checkpoints. Overall, we train 15 models, resulting in 105 different model pairs. Our results are averaged over all such pairs.

For the score-level analysis presented in Fig. \ref{fig:CelebA_scores}, we adopt the following procedure. First, we select a certain time-step $t\in[1,1000]$. Then, we select $N_{\text{eval}} = 1000$ training and test images and we apply Eq. ~\eqref{eq:forward} to each of them, resulting in $N_{\text{eval}}$ noisy images per split. Notice that the training points belong to dataset A, hence effectively representing test points for model B. The test points are taken from a third dataset different from both A and B. Finally, we compute the cosine similarity between the noise predictions (score estimates) of the two models when given the same noisy image as input:

\begin{equation}
    \text{sim}(\bm{x}_t) = C_S( \widehat{\boldsymbol{\epsilon}}_{\theta_A}(\bm{x}_t), \widehat{\boldsymbol{\epsilon}}_{\theta_B}(\bm{x}_t) ).
\end{equation}

We compute the above quantity for each of the $N_{\text{eval}}$ noisy images and we take the average. We repeat this procedure for all available training checkpoints. The final result shown in Fig. \ref{fig:CelebA_scores} is the mean (with corresponding standard error) computed over all model pairs. We present the results obtained for times $t=100, 200$ in Fig. \ref{fig:CelebA_scores_extratimes}, showing the same phenomenology as in Fig. \ref{fig:CelebA_scores}.

\subsection{Sample-level experiments}
For the sample-level analysis, we adopt the same sample-split approach as in the score-level analysis. Given a pair of models A and B, for each of them we generate $N_{\text{eval}} = 512$ images starting from random noise. Importantly, following \citep{kadkhodaie2023generalization, favero2025bigger}, we fix both the starting random noise and the reverse process noise realizations $\bm{z}$ in Eq.~\eqref{eq: DDPM reverse noise predict}. We then compute the cosine similarity between the standard-normalized final images produced by the models as: 

\begin{equation}\label{eq: cos_sim_samples}
    \text{sim}(\hat{\boldsymbol{x}}_0^A,\hat{\boldsymbol{x}}_0^B) = C_S( \hat{\boldsymbol{x}}_0^A, \hat{\boldsymbol{x}}_0^B).
\end{equation}

We average this quantity over $N_{\text{eval}}$ generated images per model and we repeat this procedure across checkpoints. Finally we compute the mean and standard error across all 105 different model pairs, resulting in Fig. \ref{fig:summary}(a).

For each model pair and for each checkpoint, we hence obtain an empirical distribution of cosine similarities built from the $N_{\text{eval}}$ pairs of generated images. In order to select the images reported in the right panel of Fig. \ref{fig:summary} (a), we consider the top 20 pairs whose cosine similarity is closer in absolute value to the mean of the aforementioned distribution (see Fig. \ref{fig:celebAtop20gens}). The images in Fig. \ref{fig:summary}(a) are extracted from such 20 pairs at different checkpoints. The closest training images to the side of each generated sample are obtained through the same approach as in Eq.~\eqref{eq: cos_sim_samples}.

\section{Additional details on controlled experiments on hierarchical data}\label{appendix:rhm_exps}

We provide the source code used to perform our numerical experiments on the repository accessible at \url{https://github.com/davide-beltrame/biased-generalization}. It includes a Python script generating the hierarchical data, as well as the PyTorch implementation of our numerical experiments, as well as an efficient implementation of the various BP denoising.

\subsection{Data model and training}

\paragraph{Dataset generation.}
Following \citet{garnier2024transformers}, we generate discrete sequences, $\bm{s}^\mu \in \mathcal{X}=\{1,\dots,q\}^N$ through a tree-based graphical model, specified by a transition tensor $\SF{M}\in\mathbb{R}_+^{q\times q\times q}$, known as the ``grammar''. The grammar assigns probabilities $M_{abc}$ to all possible production rules $a\to bc$. The generation process is then repeated over $\ell$ layers, leading to sequences of size $N = 2^\ell$. This framework is closely linked to the random hierarchy model \citep{cagnetta2024deep}, and in general to context-free grammars traditionally studied in the context of natural language processing \citep{zhao2023transformers,allen2023physics}. In our study, we focus on grammars with distinct production rules, $\mathbf{M}_a \in \mathbb{R}_+^{q\times q}$, for each ancestor symbol $a$, such that $M_{abc} M_{a'bc} = 0 \, \forall a'\neq a$. This choice ensures that the data-generating process is unambiguous, i.e., that ancestor reconstruction from an entire sequence is deterministic. Moreover, we induce sparsity by fixing the number of non-zero elements in $\mathbf{M}_a$ to $\tilde{q} < q$. The non-zero probabilities are drawn from a log-normal distribution of log-mean 0 and log-scale $\sigma$ before being normalized.

In the results presented above, we use a vocabulary size $q=6$, effective $\tilde{q} = 4$ and tree depth $\ell=4$, yielding sequence length $N=16$. 
We report averages over multiple independent training runs and data (e.g. 15 runs for the $n=12\mathrm{k}$ curves in Fig.~\ref{fig:Hierarchical_bias_evidence}), always generated with the same fixed grammar $\mathsf{M}$. The exact sequences we used in the experiments shown in this paper may be reproduced by setting \texttt{seed=0} in our data generation script.

\paragraph{Hierarchical filtering.}
To selectively suppress long-range correlations, \citet{garnier2024transformers} introduces a \emph{filtering parameter} $k \in {0,\dots,\ell}$, which truncates the hierarchy above level $k$. Nodes at depth $k$ are sampled independently conditioned on the root, while the standard branching process applies below that level.
Concretely, for each node $x_j$ at depth $k$,
\begin{equation}
\mathbb{P}(x_j = b \mid x_0 = a)
= \bigl(M_{\sigma_0(j)} M_{\sigma_1(j)} \cdots M_{\sigma_{k-1}(j)}\bigr)_{a b},
\end{equation}
where $\sigma_m(j) \in {L,R}$ denotes the branch taken at level $m$, and the effective transition matrices
\begin{equation}
(M_L)_{ab} = \sum_c M_{a b c}, \qquad
(M_R)_{ac} = \sum_b M_{a b c}
\end{equation}
are obtained by marginalizing over one child.
This construction preserves all marginals up to scale $2^{\ell-k}$, while eliminating correlations at larger scales. The cases $k=0$ and $k=\ell$ correspond to a fully hierarchical model and a conditionally independent (naive Bayes) model, respectively. In the diffusion setting of the main text, filtered models define \emph{coarse-grained oracles} that resolve only a subset of hierarchical features, and play a central role in diagnosing which structural scales are captured by a trained denoiser.

\paragraph{Continuous diffusion for discrete sequences.}
We follow the setup proposed in \citet{li2022diffusion}: each token is one-hot encoded,  $\bm{x}_{0,i}=\bm{e}_{s_i}\in\{0,1\}^q$. We use Gaussian forward diffusion (Eq.~\eqref{eq:forward}) with $T=500$ steps and we use the schedule $\overline{\alpha}_t = \prod_{i=1}^t (1-\beta_i)$ with a linear $\beta$ schedule from $\beta_1=2\times 10^{-4}$ to $\beta_T=4\times 10^{-2}$.

\paragraph{Transformer architecture and training.}
We use a transformer denoiser trained with the discrete DSM objective (cross-entropy between the true token $s_i$ and the predicted posterior mean at that position) with random timestep $t\sim\mathrm{Unif}(\{1,\dots,T\})$. The transformer has 8 layers, hidden dimension 512, 4 attention heads, and uses standard positional plus sinusoidal time embeddings. Training is performed with the Adam \citep{kingma2017adam} optimizer with learning rate $3\times 10^{-4}$, batch size 512, cross-entropy loss. Models trained on 5000, 12000 and 70000 training points are optimized for 20000, 30000 and 35000 epochs respectively. 

\subsection{Controlled oracles: Belief Propagation}
Belief Propagation is a dynamic programming algorithm that relies on message-passing along the edges of the tree to compute posterior distributions for the symbols in the graph, given knowledge of the transition tensor $\mathsf{M}$, and of a prior on the values of the symbols. Note that it is here equivalent to a simplified inside–outside algorithm for a fixed topology tree. BP proceeds by passing messages along tree edges. For a node $u$, messages are probability vectors over $\mathcal{X}$. In the unfiltered part of the tree, factor-to-variable updates read
\begin{equation}
\hat\nu_{u \to p}(x_p)
\propto
\sum_{x_u, x_{u'}} M_{x_p x_u x_{u'}}
,\nu_{u \to \alpha}(x_u),\nu_{u' \to \alpha}(x_{u'}),
\end{equation}
while variable-to-factor messages are products of incoming messages. Above the filter level, messages involve only unary conditionals $\mathbb{P}(x_j \mid x_0)$ \citep{garnier2024transformers}. Given the tree structure of the factor graph, convergence of BP is achieved in a single upward-downward pass. Once converged, the marginal of any node $i$ is given by
\begin{equation}
\mu_i(x_i) \propto \prod_{\alpha \in \partial i} \hat\nu_{\alpha \to i}(x_i).
\end{equation}

In the diffusion experiments of the main text, BP is used to compute the exact conditional posterior mean $\mathbb{E}[x_0 \mid x_t]$ under the full model and its filtered counterparts BP$_k$.
The prior, conditioning on the observation $\bm{x}_t$ is introduced in the form of a ``field'' acting on the sequence elements in the direction of a noisy observation,
\begin{equation} \label{eq:field_conditioning}
\bm{h}_t = \softmax_q\left(\frac{\sqrt{\overline{\alpha}_t }}{1-\overline{\alpha}_t} \bm{x}_t\right).
\end{equation}
Here $\frac{\sqrt{\overline{\alpha}_t }}{1-\overline{\alpha}_t}$ is the signal-to-noise ratio, recovering the setup of \cite{sclocchi2025phase}. At short times, this quantity diverges and the field pins the symbols to the value associated to the largest entry in $\bm{x}_t$. At long times, on the other hand, the signal to noise ratio will be close to zero, leading to an input that is uniform over all possible symbols, inducing BP to output the marginal probabilities $\hat{\bm{x}}_0(\bm{x}_T) = \mathbb{E}\left[ \bm{x}_0 \right]$, where the expectation is obtained from the distribution $P_0$ adapted to the one-hot encoding.


\subsection{Score-level experiments}
The score-level experiments on our data model (see Fig.~\ref{fig:summary}(b) and Fig.~\ref{fig:sequential}) follow a pipeline similar to the score-level experiments on CelebA, with one key difference. In addition to comparing the scores of models trained on different datasets (sample-split analysis), we also compare each model score against the score of each filtered BP level, \(\mathrm{BP}_k\). Concretely, we compute \(\DKL(\cdot \mid\mid \hat{\bm{x}}_0^\theta)\) for multiple choices of \(\mathrm{BP}_k\). As before, we apply Eq.~\eqref{eq:forward} to each of the \(N_{\text{eval}}=3000\) clean samples, using \(N_{\text{reps}}=5\) independent noise realizations per sample. The final results are obtained by first averaging over all noisy samples and then averaging over all model pairs formed from 15 models trained on different datasets.

\subsection{Sample-level experiments}
We conduct two types of sample-level experiments: (i) experiments based on nearest-neighbor divergence (Fig.~\ref{fig:Hierarchical_bias_evidence}(a)), and (ii) experiments using the U-turn approach (Fig.~\ref{fig:model_uturn}).

For the nearest-neighbor-divergence experiments, we proceed as follows. For a model trained on a dataset of \(n\) samples with multiple available training checkpoints, we generate \(N_{\text{eval}}\) sequences at each checkpoint. For each generated sequence, we identify its closest counterpart in the training set, defined as the training sequence with the highest overlap---i.e. the highest fraction of agreeing sequence elements, equal to 1 minus the distance normalized by the dimension. This yields an empirical distribution of maximal overlaps with the training data. We then compare it to an analogous distribution computed by, for each of \(N_{\text{eval}}\) held-out test sequences drawn from the same underlying data distribution, retrieving the nearest training sequence. Finally, for each  checkpoint, we compute the KL divergence between these two distributions. Fig.~\ref{fig:Hierarchical_bias_evidence}(a) is obtained by setting $N_{\text{eval}} = 50000$ and by averaging the curves obtained for 15 models trained on the same dataset to reduce fluctuations. Figs. \ref{fig:extra_NN_and_uturn_n12k}, \ref{fig:extra_NN_and_uturn_n5k} (left) represent the evolution of the KL divergence for two distinct models trained on 12000 and 5000 data points respectively.

For the U-turn approach, we use the following procedure. We first define a grid of diffusion time steps \(t_1,\ldots,t_K\) with \(t_i \in [1,500]\) and \(t_{i-1} < t_i\). We then select \(N_{\text{eval}} = 100\) samples from both the training and test sets. For each sample and each \(t_i\), we apply Eq.~\eqref{eq:forward} to generate \(N_{\text{reps}}=1000\) independent noised versions. Given these noised inputs and a model at a chosen checkpoint, we perform U-turns by running the reverse diffusion process starting from the noised samples. For each \(t_i\), this yields \(N_{\text{eval}} \times N_{\text{reps}}\) generated sequences. For every generated sequence, we compute its overlap with the corresponding original (clean) sequence. We then average these overlaps over \(N_{\text{reps}}\), obtaining an overlap vector of length \(N_{\text{eval}}\). We repeat the same procedure using the BP denoiser in place of the model. Finally, we take the element-wise ratio between the model and BP overlap vectors and average this ratio over \(N_{\text{eval}}\). We repeat this pipeline for all \(t_i\) and across multiple model checkpoints. Figures~\ref{fig:extra_NN_and_uturn_n12k} and \ref{fig:extra_NN_and_uturn_n5k} (right) report the U-turn results over multiple epochs for two models trained on 12000 and 5000 samples, respectively.

\subsection{Training-free toy denoiser model}
\label{appendix:toy_model}

As mentioned in Sec.~\ref{sec:setup_generative_model}, an infinitely expressive diffusion model trained from an uninformed (random-sequence) initialization will asymptotically collapse to pure memorization, behaving as the denoiser (or score) induced by the empirical data distribution.
For discrete-valued sequences, the empirical denoiser can be expressed in a BP / factor-graph formalism by replacing the binary-tree structure of the data model with a single factor
\begin{equation}
\phi(\bm{s}) = \sum_{\mu=1}^n \prod_{i=1}^N \delta_{s_i,s^\mu_i},
\end{equation}
i.e., a hard constraint that concentrates the (normalized) probability measure on the training set $\{\bm{s}^\mu\}_{\mu=1}^n$.
An architecture-free toy model of the progressive alignment with the empirical denoiser is obtained by considering a \emph{smoothing} of this factor:
\begin{equation}
\phi_\varepsilon(\bm{s}) = \sum_{\mu=1}^n \prod_{i=1}^N
\left(
\delta_{s_i,s^\mu_i}
+ \e^{-\varepsilon}\bigl(1-\delta_{s_i,s^\mu_i}\bigr)
\right),
\end{equation}
which allows deviations from the training sequences, with mismatches exponentially suppressed.
The parameter $\varepsilon$ controls the sharpness of the distribution and thus the typical distance from the training points at which samples concentrate.
After normalization, $\varepsilon=0$ yields a flat distribution over sequences, while the limit $\varepsilon\to\infty$ recovers the empirical distribution.

What is missing from this simple model is the inductive bias of the trained architecture: the $\varepsilon$-regularization is completely agnostic to the data model, and therefore the interpolation of $P_0$ far from the training points is expected to be poor.

Although one can compute the conditional posterior mean given an observation, as done in Eq.~\eqref{eq:field_conditioning}, sampling from the smoothed distribution does not require an iterative denoising diffusion mechanism.
Indeed, one may simply draw an index $\mu$ uniformly from $\{1,\ldots,n\}$—each mixture component having equal total mass by symmetry—and then sample $\bm{s}$ from the corresponding product kernel centered at $\bm{s}^\mu$.

\section{Supplementary figures}
\label{appendix:extra_figs}

\subsection{CelebA images}
\label{appendix:extra_figs_celebA}

\begin{figure}
    \centering
    \includegraphics[width=\linewidth]{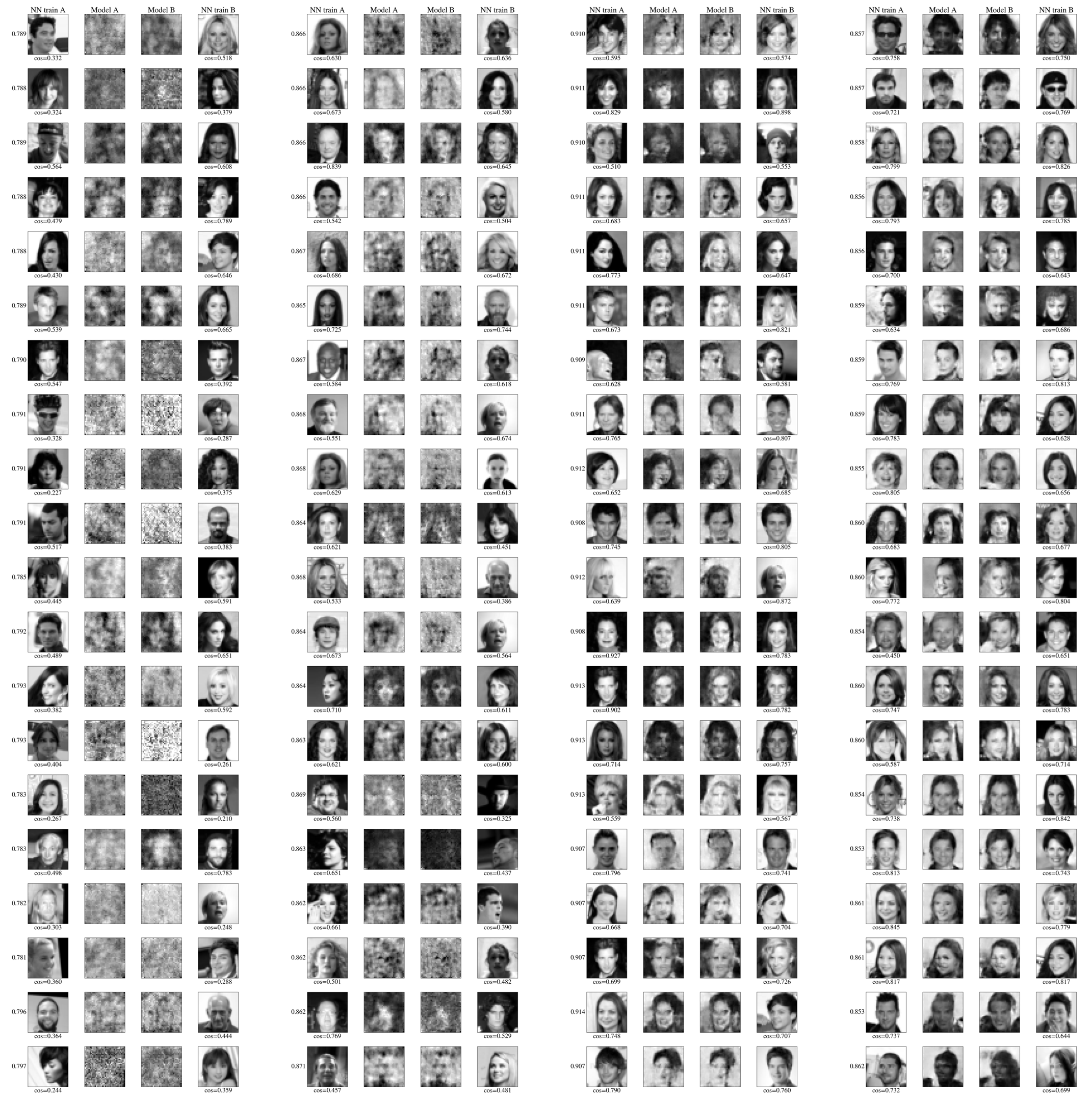}
    \caption{Set of the 20 images among 512 with the closest to typical cosine similarities between scores after (from left to right) 111, 213, 620 and 1333 epochs. As in Fig.~\ref{fig:summary}(a) (fourth row from the bottom here), central columns show samples generated by two models given the same noise trajectories, while outer columns show the closest image in their respective training sets. On the left side of each row is displayed the cosine similarity between the generated images, while below outer images is displayed the cosine similarity between the generated images and the associated closest training sample.}
    \label{fig:celebAtop20gens}
\end{figure}

\paragraph{Typical sample-split generated images.} In Fig.\,\ref{fig:celebAtop20gens}, we show sets of 20 images generated by two models trained according to the sample-split procedure described in \ref{appendix:celeba_exps}, for the same checkpoints shown in Fig.\,\ref{fig:summary}. On the left, close to initializations, the models output different yet already similar noisy samples. By the epoch shown on the center left, both models have improved significantly and get closer together. At the typical minimum of the cosine distance among generated images, essentially all rows are near identical. At the typical minimum of the test loss, some visible discrepancies start arising in some (albeit not all) of the generations.

\begin{figure}
    \centering
    \includegraphics[width=0.5\linewidth]{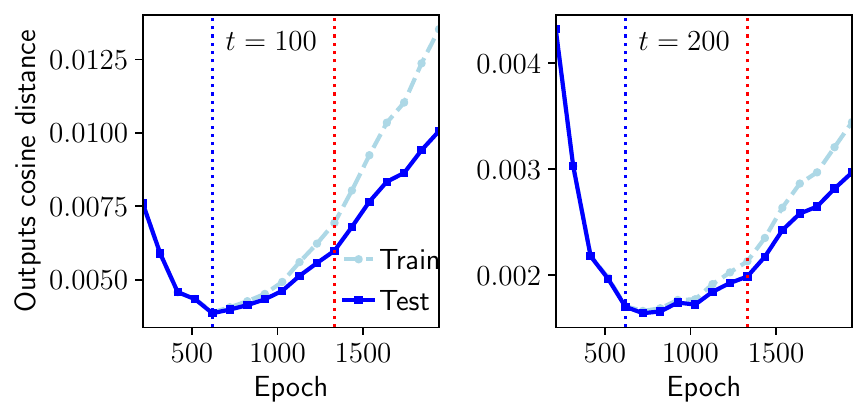}
    \caption{Reproduction of the experiment presented in Fig.~\ref{fig:CelebA_scores} for two additional fixed diffusion times.}
    \label{fig:CelebA_scores_extratimes}
\end{figure}

\paragraph{Sample-split comparison of models at fixed times.} In Fig.~\ref{fig:CelebA_scores_extratimes}, we show additional diffusion times for the comparison of models trained in CelebA, as in Fig.~\ref{fig:CelebA_scores}. The phenomenology is identical to what is described in Sec.\,\ref{sec:biased_CelebA}. However, for longer times, we see that the impact of biased generalization is less pronounced in the denoiser's behavior, as expected from the vanishing data-dependence of the long-time regime of diffusion.

\subsection{Hierarchical data model}
\label{appendix:extra_figs_hierarchical}

\begin{figure*}
    \centering
    \includegraphics[width=\linewidth]{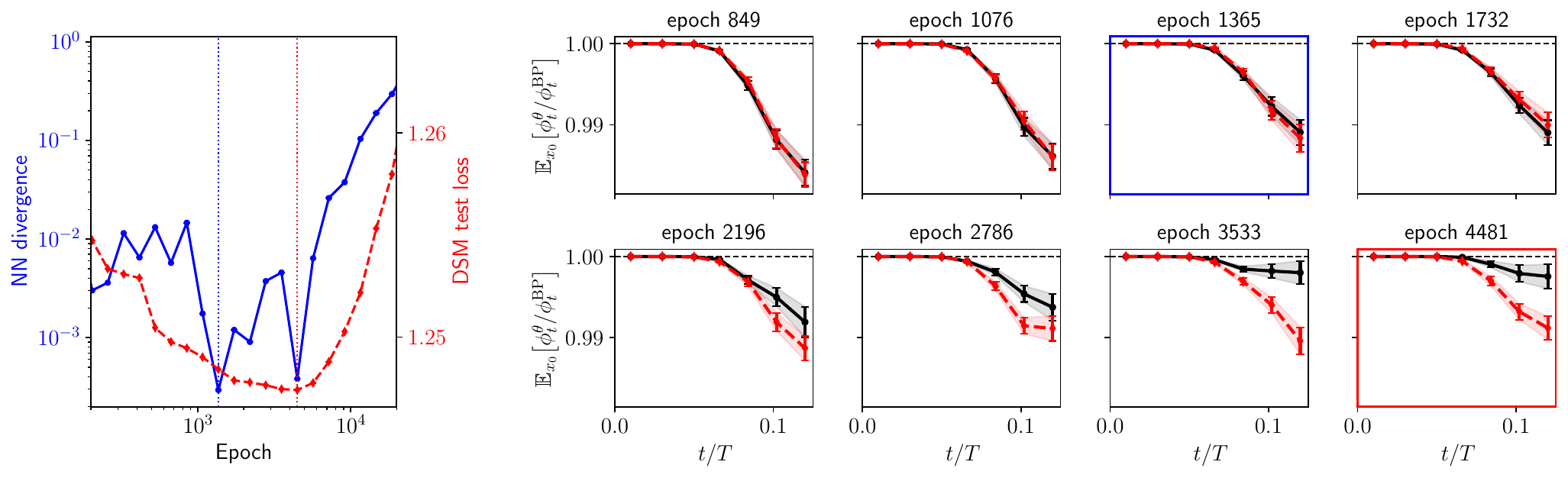}
    \caption{Left: NN divergence and DSM test loss versus training epoch for a model trained on $n=12\mathrm{k}$ samples of hierarchical data, with respective minima indicated by vertical dotted lines. Right: U‑turn overlap ratios across epochs for starting points in the training set of the model (black lines) or from a test set (red dashed lines). Blue/red framed panels highlight the epochs corresponding to the bias metric minimum and test‑loss minimum respectively (shown in Fig.~\ref{fig:model_uturn}).}
    \label{fig:extra_NN_and_uturn_n12k}
\end{figure*}

\begin{figure*}
    \centering
    \includegraphics[width=\linewidth]{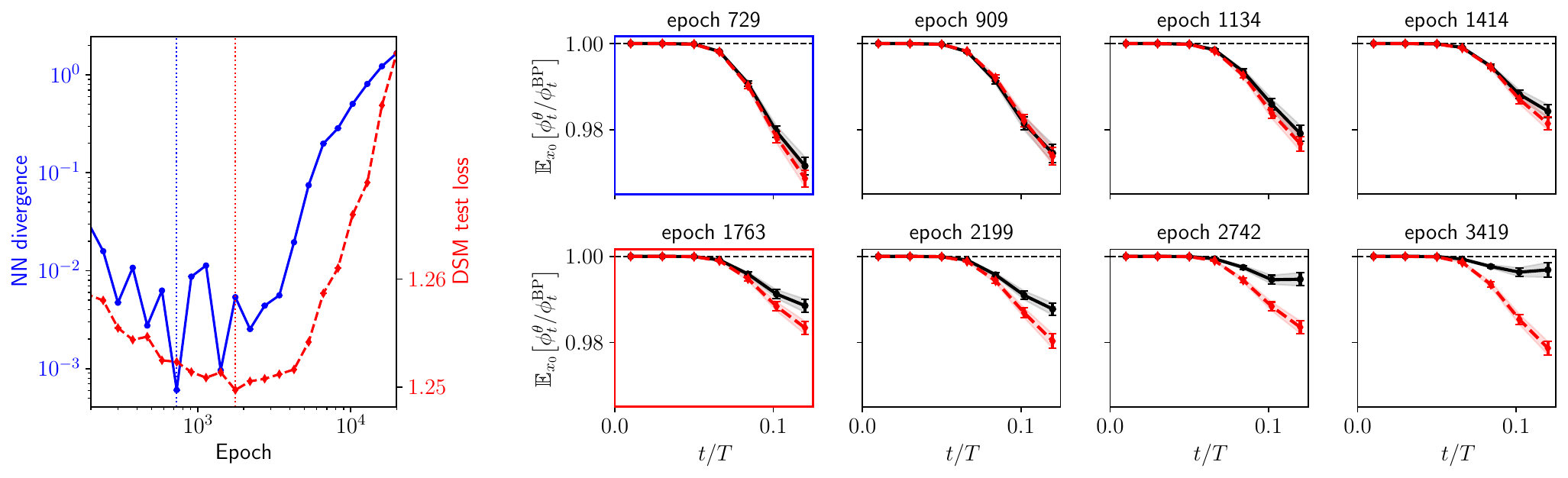}
    \caption{Left: NN divergence and DSM test loss versus training epoch for a model trained on $n=5\mathrm{k}$ samples of hierarchical data, with respective minima indicated by vertical dotted lines. Right: u‑turn overlap ratios across epochs for starting points in the training set of the model (black lines) or from a test set (red dashed lines). Blue/red framed panels highlight the epochs corresponding to the bias metric minimum and test‑loss minimum respectively.}
    \label{fig:extra_NN_and_uturn_n5k}
\end{figure*}

\paragraph{Single-instance NN divergence.} In Fig.~\ref{fig:extra_NN_and_uturn_n12k} (left) and Fig.~\ref{fig:extra_NN_and_uturn_n5k} (left), we show two representative seeds of the experiment shown in Fig.\,\ref{fig:Hierarchical_bias_evidence} (left), where the curves are averaged over $15$ independent runs. As can be seen from these curves, the Adam optimization dynamics introduces strong oscillations that significantly affect the nearest-neighbor divergence. While the described phenomenon is robust across runs, the amount of bias recorded at the minimum of the test loss---that can be quantified by looking at the difference between the two values of the blue curve in correspondence of the dashed lines, is training trajectory dependent and can vary from run to run. As also shown in these figures and described just below, other methods of diagnostic such as the U-turn experiments may thus be more robust than this extreme value statistic for single instances.

\paragraph{Full sequence of U-turns across training.}
Figures~\ref{fig:extra_NN_and_uturn_n12k} (right) and \ref{fig:extra_NN_and_uturn_n5k} (right) report the U-turn results on our data for two models trained on 12{,}000 and 5{,}000 training samples, respectively. We find that, at the epoch corresponding to the minimum test loss, the models are closer to BP, even though they exhibit asymmetric behavior between training and test data. In contrast, at earlier epochs the models behave more similarly across the two splits, but this comes at the expense of poorer generalization.

\paragraph{U-turns on random starting sequences.} We further investigate the behavior of our trained denoisers in the short-time dynamical regime (regime (iii) in Sec.\,\ref{sec:exact_phenomenology}). In particular, in the main, we described this regime as \emph{trivial}, since all our oracle denoisers and the trained models display a negligible score divergence $D_\mathrm{KL}$ for $t/T\lesssim0.8$, provided the comparison is performed around valid i.i.d. datapoints $\bm{x}_0$. Note that the equivalence class could be further extended to the empirical score, if the sequences were selected from the training set.

However, we also argued that this similarity would break if we considered a reference $\bm{x}_0$ that has zero probability in some of the models, and non-zero in others. 
To investigate this, we follow similar lines to the U-turn experiment described in Sec.\,\ref{sec:biased_models}. As an extreme example, we consider reference sequences sampled uniformly at random from $\{1,..,q\}^N$, and their corresponding one-hot encodings $\bm{x}_0$. What we expect is that a U-turn process starting from noised random sequences could come back to the original point only when it has a non-vanishing probability in the associated posterior measure. Therefore, more selective scores (in order: the empirical score, BP$_0$, BP$_1$, ...) should obtain a smaller average overlap $\mathbb{E}[\phi_t(\bm{x}_0)]$. 

\begin{figure}
    \centering
    \includegraphics[width=0.5\linewidth]{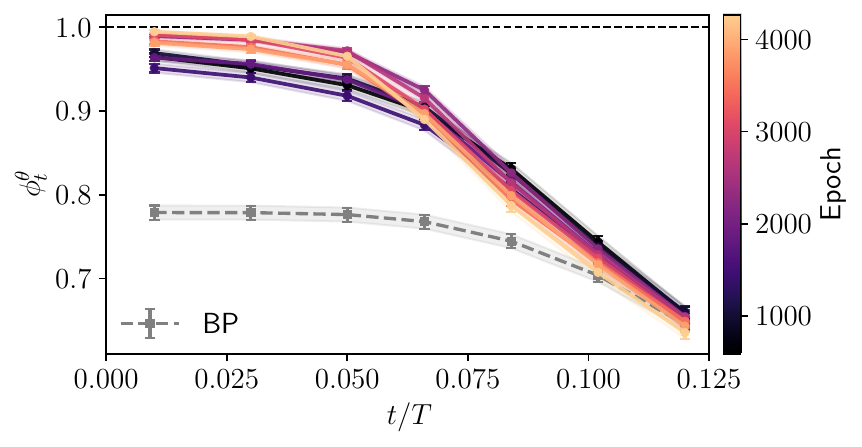}
    \caption{U-turn experiment performed by noising sequences whose discrete elements are drawn uniformly at random. The left y-axis shows the averaged overlap of the generated sequences with the original ones.}
    \label{fig:Uturn_randomstart}
\end{figure}

In Fig.\,\ref{fig:Uturn_randomstart}, we show the typical overlap achieved with the BP oracle score (grey dashed line), and the corresponding overlap obtained with our trained model (different epochs in different colors) from this experiment. 
The results confirm that the trained model very quickly learns a simpler explanation, here likely rounding up to the closest $\{0,1\}$ sequence to obtain a one-hot vector, instead of the ground truth. 
This finding confirms that the trained denoiser is approximating the behavior of BP and the empirical score ``by coincidence'' in the trivial phase, since the learned short-time function is compatible with the former denoisers only in proximity to specific points in input space.

\end{document}